\documentclass[10pt,twocolumn,letterpaper]{article}

\usepackage{wacv}              

\usepackage{graphicx}
\usepackage{amsmath}
\usepackage{amssymb}
\usepackage{booktabs}
\usepackage{fontawesome}
\usepackage{multicol}
\usepackage{multirow}
\usepackage{enumitem}
\usepackage{breakcites}
\usepackage[dvipsnames]{xcolor}

\usepackage{algorithm}
\usepackage{algorithmicx}
\usepackage{algpseudocode}
\usepackage{dsfont}
\usepackage{xcolor}

\newcommand{\blockcomment}[1]{}

\DeclareMathOperator*{\argmax}{argmax}

\newcommand{\nb}{\nabla}
\newcommand{\T}{\top}

\newcommand{\R}{\mathbb{R}}

\newcommand{\cA}{\mathcal{A}}

\newcommand{\cD}{\mathcal{D}}
\newcommand{\cE}{\mathcal{E}}

\newcommand{\cG}{\mathcal{G}}

\newcommand{\cL}{\mathcal{L}}
\newcommand{\cM}{\mathcal{M}}
\newcommand{\cN}{\mathcal{N}}

\newcommand{\cS}{\mathcal{S}}
\newcommand{\cT}{\mathcal{T}}
\newcommand{\cX}{\mathcal{X}}
\newcommand{\cY}{\mathcal{Y}}

\newcommand{\bx}{\mathbf{x}}
\newcommand{\bu}{\mathbf{u}}
\newcommand{\bI}{\mathbf{I}}

\renewcommand{\a}{\alpha}

\newcommand{\e}{\varepsilon}

\newcommand{\s}{\sigma}

\usepackage[pagebackref,breaklinks,colorlinks]{hyperref}

\usepackage[capitalize]{cleveref}
\crefname{section}{Sec.}{Secs.}
\Crefname{section}{Section}{Sections}
\Crefname{table}{Table}{Tables}
\crefname{table}{Tab.}{Tabs.}

\def\wacvPaperID{223} 
\def\confName{WACV}
\def\confYear{2025}

\begin{document}

\title{Domain-Guided Weight Modulation for Semi-Supervised Domain Generalization}

\author{Chamuditha Jayanaga Galappaththige$^{1}$ \hspace{15pt} Zachary Izzo$^{2}$\hspace{15pt} Xilin He$^{3}$ \hspace{15pt} Honglu Zhou$^{4}$ \\ Muhammad Haris Khan$^{1}$ \\
$^1$MBZUAI, UAE \hspace{6pt}  $^2$NEC Labs, USA  \hspace{6pt} $^3$Shenzhen University, China \hspace{6pt} $^4$Salesforce AI Research, USA \\
}
\maketitle

\begin{figure*}[t]
  \centering
   \includegraphics[width=1.0\linewidth]{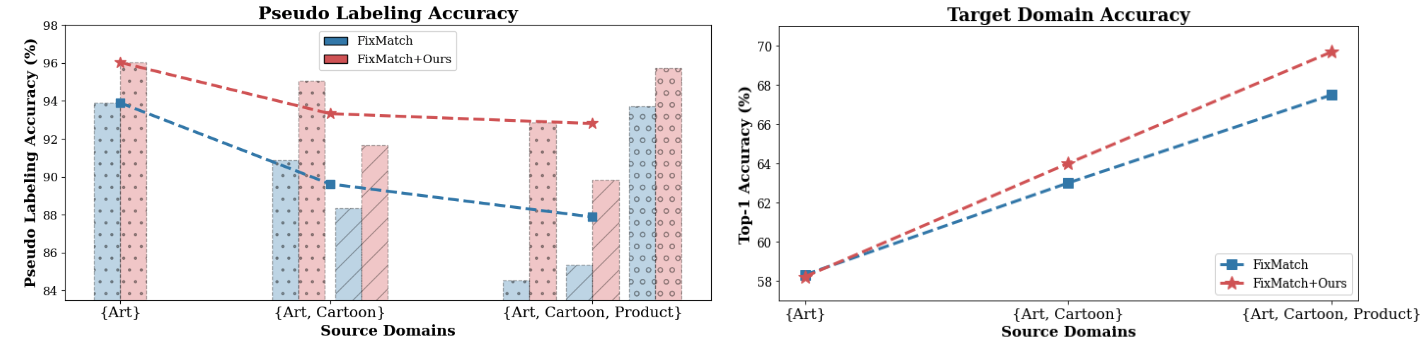}
   \caption{Left: Pseudo-labeling (PL) accuracy when source domains (Art, Cartoon, and Product) are gradually added to the set of training domains in baseline \cite{sohn2020fixmatch} and ours on OfficHome dataset. Our method tends to maintain a higher PL accuracy while the baseline's PL accuracy drops upon gradually adding source domains. Right: Top-1 Accuracy on the target domain (Real-world). }
   
   \label{fig:PL_accuracy_adding_source_domains}
\end{figure*}
\begin{abstract}
   Unarguably, deep learning models capable of generalizing to unseen domain data while leveraging a few labels are of great practical significance due to low developmental costs. In search of this endeavor, we study the challenging problem of semi-supervised domain generalization (SSDG), where the goal is to learn a domain-generalizable model while using only a small fraction of labeled data and a relatively large fraction of unlabeled data. Domain generalization (DG) methods show subpar performance under the SSDG setting, whereas semi-supervised learning (SSL) methods demonstrate relatively better performance, however, they are considerably poor compared to the fully-supervised DG methods. Towards handling this new, but challenging problem of SSDG, we propose a novel method that can facilitate the generation of accurate pseudo-labels under various domain shifts. This is accomplished by retaining the domain-level specialism in the classifier during training corresponding to each source domain. Specifically, we first create domain-level information vectors on the fly which are then utilized to learn a domain-aware mask for modulating the classifier's weights. We provide a mathematical interpretation for the effect of this modulation procedure on both pseudo-labeling and model training. Our method is plug-and-play and can be readily applied to different SSL baselines for SSDG. Extensive experiments on six challenging datasets in two different SSDG settings show that our method provides visible gains over the various strong SSL-based SSDG baselines. Our code is available at \href{https://github.com/Chumsy0725/DGWM}{github.com/Chumsy0725/DGWM}.
\end{abstract}

\section{Introduction}
\label{sec:intro}

\noindent\textbf{Background:} The problem of domain shift, which violates the i.i.d. assumption of data between the training and testing distributions, causes top-performing visual recognition models \cite{he2016deep, dosovitskiy2020image} to (substantially) lose performance \cite{tzeng2014deep, li2017deeper, hoffman2018cycada, recht2019imagenet, sultana2022self}. To tackle this problem, the research direction of domain generalization (DG) has received great attention in the recent past \cite{li2019episodic, carlucci2019domain, khan2021mode, sultana2022self}. The conventional DG setting assumes that the data from multiple source domains are available and the aim is to train a model that can show adequate precision in the data from an unseen target domain \cite{blanchard2011generalizing, muandet2013domain, li2017deeper}. The DG problem has been tackled from different directions, consequently, we have witnessed promising progress so far. A myriad of DG approaches have been proposed which, for instance, employ auxiliary tasks \cite{carlucci2019domain,wang2020learning}, diversify source domains \cite{zhou2020learning, khan2021mode}, or simulate DG scenario while training \cite{li2019episodic}. However, these (traditional) DG methods are developed on the assumption that the data from multiple source domains are fully labeled.

\noindent\textbf{SSDG settings:} We study the problem of semi-supervised domain generalization (SSDG), which combines the problems of domain generalization and label-efficient learning \cite{zhou2023semi, galappaththige2024towards}. SSDG could serve many real-world applications, e.g., autonomous drone navigation, where acquiring a large set of labeled data can be expensive, time-consuming, or even infeasible. SSDG and DG are similar in task-level objective which expects learning a domain-generalizable model from different source domains \cite{zhou2023semi, galappaththige2024towards}. As mentioned earlier, the DG setting assumes that the data from source domains is completely labeled. However, the SSDG setting is based on semi-supervised learning (SSL), where only a small portion of labeled data is provided and the majority of data is unlabeled \cite{yuan2022label}. DG methods tend to struggle under the SSDG setting, primarily because they are not designed to leverage unlabeled data. On the contrary, SSL methods display relatively better performance under SSDG setting, however, they are notably inferior to fully-supervised DG methods \cite{zhou2023semi}. SSL-based SSDG methods commonly use a domain-shared classifier to pseudo-label the unlabeled data. A domain-shared classifier is a single classifier shared among all source domains during training.

\noindent\textbf{Our motivation:} We propose a new SSDG approach after identifying the key limitations in the top-performing SSL-based SSDG baselines. Through preliminary investigation, we observe that the pseudo-labeling (PL) accuracy of SSL-based SSDG baselines begins to drop upon adding training data from multiple source domains (see Fig.~\ref{fig:PL_accuracy_adding_source_domains}). This is likely because the domain-shared classifier tends to sacrifice the domain-level specialism after it observes data from multiple domains having different distributions therefore hurting the PL accuracy. Poor PL accuracy during training directly affects the DG capability of the model.

\noindent\textbf{Contributions:} To this end, we propose to retain the domain-level specialism of the classifier corresponding to a particular source domain when it observes data from multiple source domains so that it can produce accurate pseudo-labels (see Fig.~\ref{fig:PL_accuracy_adding_source_domains}). This is realized by learning a domain-guided weight modulation mask which is used to modulate the weights of the (domain-shared) classifier on the fly during model training. 
Particularly, our method first curates domain-level information, then learns a mapping from this information to low-rank decomposed factors of modulation mask, which are then combined to construct a domain-guided weight modulation mask. We summarize our key contributions as follows:
\begin{itemize}
\item We explore a relatively underexplored and challenging problem of SSDG, and identify that the strong SSL-based SSDG baseline starts to lose PL accuracy upon adding data manifesting various domain shifts.
\item We propose a new approach of attaining a domain-level specialism in a classifier corresponding to each source domain by learning a domain-guided weight modulation mask to modulate the classifier's weights during training. We provide a mathematical interpretation of the effect of the weight modulation on both pseudo-labeling and model training dynamics.
\item Our approach is plug-and-play and can be readily integrated into different SSL-based SSDG baselines. Extensive experimental results on six challenging DG datasets with two different SSDG settings show that our method provides notable gains over different baselines under different distribution shifts.
\end{itemize}

\section{Related work}

\noindent\textbf{Domain Generalization:}
The objective of domain generalization(DG) is to learn robust representations that are independent of domain-specific factors and thus can generalize well to the unseen target domains. Existing methods can be substantially categorized into domain alignment \cite{domain_alignment1, domain_alignment2, domain_alignment3}, data augmentation \cite{data_augmentation1, data_augmentation2, data_augmentation3} and meta-learning \cite{meta_learning1, meta_learning2, meta_learning3}. Domain alignment techniques \cite{domain_alignment1, domain_alignment2, data_augmentation3} strive to cultivate a domain-agnostic feature space by mapping samples from multiple domains into a unified subspace. Meanwhile, data augmentation methods \cite{data_augmentation1, data_augmentation2, data_augmentation3} tend to generate virtual data, which serves to boost the data diversity. On the other hand, meta-learning methods \cite{meta_learning1, meta_learning2, meta_learning3} construct episodes by partitioning the source domains into non-overlapping meta-train and meta-test sets and strike to train a model with improved performance on the meta-test sets. However, a significant limitation looms over these methodologies as a majority of the existing domain generalization techniques are ill-equipped to process unlabeled data, largely stemming from their foundational assumption of a fully supervised learning context.

\noindent\textbf{Semi-Supervised Learning:} Existing work on semi-supervised learning mainly consists of consistency regularization, entropy minimization, and pseudo-labeling. Consistency learning methods \cite{miyato2018virtual, tarvainen2017mean, xie2020self, sohn2020fixmatch, ouali2020semi} operate on the principle of a classification model should favor function that produces consistent outputs for similar data instances which minimizes the cost on a manifold around each data instance \cite{tarvainen2017mean}. Consistency can be achieved by adding noise to the inputs \cite{berthelot2019mixmatch,sohn2020fixmatch,xie2020unsupervised, miyato2018virtual}, adding noise to the model \cite{dropout,NIPS2014_66be31e4,10.5555/3157096.3157227, laine2017temporal}, imposing consistency loss on penultimate features \cite{abuduweili2021adaptive} or on model outputs \cite{sohn2020fixmatch, tarvainen2017mean}. Entropy minimization \cite{grandvalet2004semi} enforces a classifier to output low entropy predictions on unlabeled instances using an objective function that minimizes the entropy of model prediction given an unlabeled instance. Pseudo Label \cite{lee2013pseudo} implicitly achieves entropy minimization by constructing either a hard or soft artificial label from a high-confidence prediction on an unlabeled instance using a model under training \cite{sohn2020fixmatch} or a pre-trained model \cite{xie2020unsupervised}. Recently, a line of work \cite{wang2023freematch,chen2023softmatch,zhang2021flexmatch} has been proposed building upon FixMatch \cite{sohn2020fixmatch}. \cite{wang2023freematch} propose a method to adjust FixMatch's threshold in a self-adaptive manner. \cite{chen2023softmatch} introduces a soft version of thresholding to FixMatch while \cite{zhang2021flexmatch} boost FixMatch's performance with curriculum labelling. For the SSDG problem, SSL methods, such as FixMatch \cite{sohn2020fixmatch}, tend to show more encouraging performance than DG methods.  

\noindent\textbf{Semi-Supervised Domain Generalization:}
Semi-supervised domain generalization has emerged as a promising avenue to address the challenges posed by domain shifts with limited labeled data \cite{yuan2022label}. However, only a few works have been proposed on SSDG, leaving it in an unexplored but more realistic direction. There are two main settings used in the SSDG literature. \cite{zhou2023semi, yeGraph, labelefficient, galappaththige2024towards} retain only a few instances of each source domain as labeled. While \cite{wang2023better} keeps one source domain fully labeled and others fully unlabelled. It's worth noting that the two settings are completely different and possess unique challenges endemic to that setting. To the best of our knowledge, ours is the first SSDG method that shows notable improvements under both settings. 

Recently, Zhou et al. \cite{zhou2023semi} introduced StyleMatch, which extends the FixMatch \cite{sohn2020fixmatch} with stochastic modeling and multi-view consistency learning to mitigate overfitting. \cite{yeGraph} proposed a graph laplacian regularizer that relies on the generated similarity graph and \cite{labelefficient} introduced a framework that jointly optimizes active exploration and semi-supervised generalization. \cite{galappaththige2024towards} introduced two losses to improve PL accuracy and regularize the feature space while \cite{multimatch} proposed multi-task learning framework considering each training domain as a local task and and combining all training data as a global task. \cite{zhang2023semisupervised} studies SSDG problem with known and unknown classes. The most related work to ours is \cite{wang2023better} which proposed a joint domain-aware label and dual-classifier. It improves pseudo-labeling by employing a separate classifier and maintaining a memory bank for each class of each domain created with high-confidence predictions from the previous epoch. 
Notably, it addresses only the second setting i.e. one source domain is fully labeled and the others fully unlabelled. Moreover, it utilizes a complicated training procedure that involves a memory bank, a separate dual classifier, a discriminator, domain mixup \cite{zhang2018mixup} guided adversarial training setup and several objective functions. Different to \cite{wang2023better}, to improve pseudo-labeling under shifts, we aggregate domain information available at a minibatch and use it to learn a soft mask to modulate the domain-shared classifier weights on-the-fly. Ours is a relatively simpler method, does not introduce any new losses or complicated training procedures, and yet shows remarkable performance gains in both settings with different baselines.

\section{Method}
\subsection{Notation \& Preliminaries}
Our notation is adapted from \cite{zhou2022survey}. 
A \emph{domain} is defined by a joint probability distribution $P_{XY}$ over the features and label space ${\cX \times \cY}$. In this work, $\cX=\R^D$ is the space of images represented as real vectors, and $\cY=\{1,\ldots,C\}$ is a set of $C$ possible classes. We assume that we have datasets $\cS^{(k)}$ drawn from $K$ \emph{source} domains $P^{(k)}_{XY}$, $k=1,\ldots,K$. In the first semi-supervised setting, each source dataset $\cS^{(k)}$ consists of both labelled and unlabelled data: $\cS^{(k)} = \cS^{(k)}_\ell \cup \cS^{(k)}_u$, with $\cS^{(k)}_\ell = \{(\bx^{(k)}_i, y^{(k)}_i)\}_{i=1}^{n_\ell}\sim_{\iid} P^{(k)}_{XY}$ and $\cS^{(k)}_u = \{\bu^{(k)}_i\}_{i=1}^{n_u}\sim_{\iid} P^{(k)}_X$, where $P^{(k)}_X$ is the marginal distribution of $P^{(k)}_{XY}$ over $\cX$. In practice, the unlabelled dataset $\cS^{(k)}_u$ will consist of both samples for which we actually do not have a label, as well as the feature vectors for labeled samples with their labels dropped. In the semi-supervised setting, we assume that there is much more unlabelled than labeled data, i.e. $n_u \gg n_\ell$. Following StyleMatch \cite{zhou2023semi}, we will have $n_\ell \in \{5, 10\}$ for the first setting. In the second SSDG setting, we keep one source domain completely labeled and the other source domains completely unlabelled.
Given this data, our goal is to produce a classifier $h$ for a \emph{target} domain $P^\cT_{XY}$, such that $h(\bx^\cT) = y^\cT$ with high probability when $(\bx^\cT, y^\cT)\sim P^\cT_{XY}$. Our learned model $h$ consists of a feature embedding $f:\cX \to \R^d$ and classifier weights $W \in \R^{C\times d}$, so that $h(\bx) = \mathrm{softmax}(Wf(\bx))$.

\subsection{Baselines, Their Limitations \& Our Motivation} 

\noindent\textbf{Baselines:} Our method is model-agnostic and plug-and-play; it can be seamlessly integrated with different SSL and SSDG approaches. We show the applicability of our method (see Sec.~\ref{secttion:Experiments}) with the following SSL approaches: Entropy minimization \cite{grandvalet2004semi}, MeanTeacher \cite{tarvainen2017mean}, FixMatch \cite{sohn2020fixmatch}, FBCSA \cite{galappaththige2024towards} and StyleMatch \cite{zhou2023semi}. Here, we choose FixMatch to explain our method since it emerged as the competitive SSDG baseline and combines both pseudo-labeling and consistency regularization techniques. Pseudo-labeling generates an artificial label for an unlabelled example if the $\arg \max$ of the model's prediction probability for the respective example is over a predefined threshold. Whereas consistency regularisation leverages unlabelled data by bringing the prediction on an unlabelled example as similar as possible to the prediction on a (strongly) perturbed version of the same unlabelled example. FixMatch processes an unlabelled image on two branches:~a weak-augmentation branch (pseudo-labelling branch) and a strong-augmentation branch (learning branch). The weak-augmentation branch constructs a pseudo-label on the weak augmented version \cite{sohn2020fixmatch} of an image which is then used as the target for the prediction corresponding to the strong augmented version \cite{sohn2020fixmatch} of the same image generated by the strong-augmented branch. A cross-entropy loss, denoted by $\mathcal{L}_u$, is used to enforce the consistency between the two views of the unlabelled image. The overall loss for the FixMatch is formulated as $\mathcal{L}=\mathcal{L}_u+\mathcal{L}_s$ where $\mathcal{L}_s$ is the cross entropy loss applied over labeled images separately. 

\noindent\textbf{Limitations and our motivation:} SSL-based SSDG baselines (e.g., FixMatch \cite{sohn2020fixmatch}) demonstrate encouraging performance in SSDG setting. However, there is still considerable room for improvement when comparing their performance to fully-supervised DG methods. Our preliminary experiments show that a core reason is consistent and notable deterioration in pseudo-labeling (PL) accuracy upon increasing source domains bearing different distribution shifts (see Fig.~\ref{fig:PL_accuracy_adding_source_domains}). This could be because the classifier tends to lose the domain-level specialism when operating on data from different distributions. Undoubtedly, a degrading PL accuracy negatively impacts the attainment of domain generalization capability.
A straightforward solution is to employ separate classifiers for each domain. We empirically show that such a naive solution does not improve SSDG performance (see Tab.~\ref{tab:ablation}) likely due to available data points being further constrained making classifiers prone to overfitting. Moreover, such a solution is not possible in the case of the second setting as only one domain is fully labeled and the rest are completely unlabelled. 

\begin{figure}[t]
  \centering
   \includegraphics[width=1.0\linewidth]{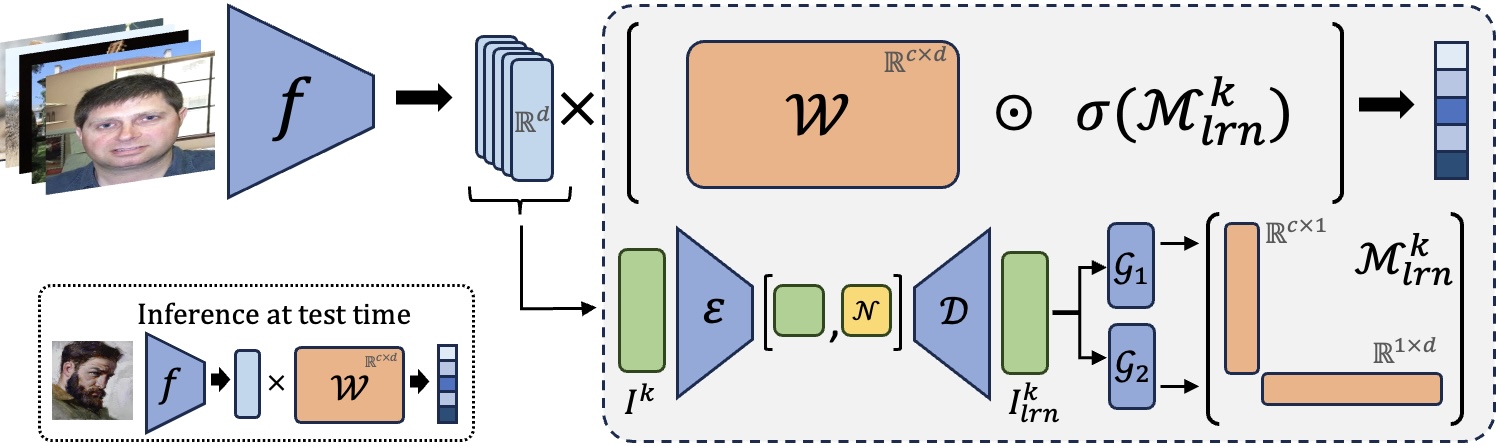}
   \caption{Overall architecture with our domain-guided weight modulation method.}
     \label{fig:overall_architecture} \vspace{-1em}
\end{figure}

To tackle this limitation, we propose to impart the domain-level specialism in the classifier corresponding to each source domain when it faces multi-source data. We actualize this by learning domain-guided weight modulation for the classifier (sec.~\ref{subsection:Domain information guided weight modulation}) to induce the domain-level specialism on-the-fly during training. Next, we develop a mathematical interpretation of the impact of our weight modulation on both pseudo-labeling and model training dynamics (sec.~\ref{subsection:Effects of Weight Modulation}). Fig.~\ref{fig:overall_architecture} displays the overall architecture with our domain-guided weight modulation method.

\subsection{Domain-Guided Weight Modulation}
\label{subsection:Domain information guided weight modulation}
We provide a complete description of our algorithm, summarized in Algorithm~\ref{alg:hyper}. Next, we discuss each component of the algorithm in detail.
During training, we process samples in minibatches consisting of samples from the same domain. We denote the index set of labeled and unlabelled examples in a minibatch by $B^{(k)}_\ell$ and $B^{(k)}_u$, respectively. The batch indices will always be defined so that $B^{(k)}_u$ also contains the feature vectors corresponding to the labeled examples in this batch, with their labels dropped (which we have included in the complete unlabelled dataset $\cS^{(k)}_u$). The superscript $(k)$ emphasizes that all of these examples are sampled from the same domain, and will be drawn from the $k$-th source dataset $\cS^{(k)}$.

\noindent\textbf{Domain information aggregation:} We would like to aggregate the domain-specific information from the minibatch and use it to eventually improve the pseudo-labeling. To do this, we compute a \emph{domain information} vector:
\begin{equation}
    \bI^{(k)} = \frac{1}{|B^{(k)}_u|} \sum_{i\in B^{(k)}_u} f(\bu^{(k)}_i).
    \label{eq:domain-info}
\end{equation}

The mini-batch mean is a simple way of aggregating domain-specific information and is also motivated by \cite{Huang2017ArbitraryST, dumoulin2017a}. We further compare different approaches for domain information aggregation in Table \ref{tab:domain_analysis}.


\noindent\textbf{Weight modulation for pseudo-labeling:}
To produce more accurate pseudo-labels across varied domains, we include a \emph{weight modulation} component which specializes the domain-shared classifier weights to the domain being processed in the current minibatch. We perform the modulation via a soft masking procedure. Specifically, we compute a matrix $\cM^{(k)}_{\mathrm{ss}} \in [0,1]^{C\times d}$ of the same shape as the (domain-shared) last-layer classifier $W$. The domain-specialized pseudo-labeling classifier is then given by $W^{(k)}_{\mathrm{ss}} = W\odot \cM^{(k)}_{\mathrm{ss}}$, where $\odot$ denotes the elementwise product. Intuitively, $\cM^{(k)}_{\mathrm{ss}}$ should downweight features that are not important for domain $k$, making it easier to generate more accurate pseudo-labels.

In order to compute the soft mask $\cM^{(k)}_{\mathrm{ss}}$, we avail ourselves of the domain information vector $\bI^{(k)}$ and let the mask be a learned function of this vector: $\cM^{(k)}_{\mathrm{ss}} = \cG(\bI^{(k)})$. Rather than learning a fully general map $\bI^{(k)}\mapsto \cM^{(k)}_{\mathrm{ss}}$, we instead enforce a special structure, as detailed below.

The transformation consists of several steps. First, we use a learned encoder/decoder-like pair $\cE: \R^d \rightarrow \R^l$ and $\cD: \R^l \rightarrow \R^d$. We map the domain information $\bI^{(k)}$ through this pair to obtain $\bI^{(k)}_{\mathrm{ss}}$. The reason for using an encoder/decoder-like pair is to inject noise into the domain information vector during the learning branch which we will further explain in the next steps.  
Next, we map the reconstructed domain information to the soft mask using a special low-rank structure. Specifically, we define two learnable transformations $\cG_1: \R^d \rightarrow \R^{C \times 1}$ and $\cG_2: \R^d \rightarrow \R^{1 \times d}$. The weight modulation matrix is then computed as
\begin{equation}
     \cM^{(k)}_{\mathrm{ss}} = \s( \cG_1(\bI^{(k)}_{\mathrm{ss}}) \times \cG_2(\bI^{(k)}_{\mathrm{ss}})), 
\end{equation}
where $\s$ is the elementwise $sigmoid$ function. 
Note that there is no specific reconstruction loss used to train the encoder/decoder-like pair; in fact, this entire step can be folded into $\cG_1$ and $\cG_2$, in effect making these deeper MLPs with shared early layer representations. The reason for considering the first two steps in the pipeline as an encoder/decoder-like pair and low-rank structure will become more clear in the next step (weight modulation for learning).

\noindent\textbf{Performing the pseudo-labeling:}
Now that we have the modulated weights, we can perform pseudo-labeling. Following FixMatch \cite{sohn2020fixmatch}, for each unlabelled example in the batch, we check whether or not the model's maximum confidence on this example is above a threshold:
\begin{equation}
     \max \mathrm{softmax}(W^{(k)}_{\mathrm{ss}} f(\a(\bu^{(k)}_i))) \geq \tau.
\end{equation}
$\a$ is a weak data augmentation function (see FixMatch \cite{sohn2020fixmatch}). For each index $i \in B^{(k)}_u$ where this inequality holds, we set the pseudo-label $\tilde{y}^{(k)}_i = \argmax W^{(k)}_{\mathrm{ss}} f(\a(\bu^{(k)}_i))$ to be the model's prediction on the weakly augmented sample, and we add this sample to a list $B^{(k)}_{\mathrm{ss}}$ of pseudo-labelled points.
This completes the pseudo-labeling branch of the algorithm. The hard thresholding used in the creation of the pseudo-label means that this branch of the computation will not give us useful gradients for learning; indeed, although the pseudo-labels $\tilde{y}^{(k)}_i$ depend on the learned components, they will be considered as constants when we perform gradient computations \cite{sohn2020fixmatch}. To obtain useful gradients and actually learn these components, we now describe the learning branch of the computation.

\noindent\textbf{Weight modulation for learning:}
In the learning branch, we inject noise into the encoder/decoder-like pair to encourage the model to learn more robust domain information. A line of work \cite{berthelot2019mixmatch, sohn2020fixmatch,xie2020self,xie2020unsupervised} has demonstrated that introducing noise to the representations can improve consistency learning. Our baseline FixMatch introduces noise to the inputs using a strong augmentation function in their learning branch \cite{sohn2020fixmatch}. In addition to that, we use a noise-injected encoder/decoder-like architecture to perturb the domain information vector $\bI^{(k)}$ and therefore introduce noise to the mask generation process $\bI^{(k)}\mapsto \cM^{(k)}_{\mathrm{lrn}}$ which will be eventually reflected in domain-specialized classifier weights $W^{(k)}_{\mathrm{lrn}} = W\odot \cM^{(k)}_{\mathrm{lrn}}$.
Thus, we compute
\begin{equation}
    \bI^{(k)}_{\mathrm{lrn}} = \cD(concat(\cE(\bI^{(k)}) , \cN(0, \e^2I))). 
\end{equation}
Here, $\cN(0, \e^2I)$ is an isotropic Gaussian of the same shape as $\cE(\bI^{(k)})$. The variance $\e^2$ is a hyperparameter. We keep $\e^2=0$ in the pseudo-labeling branch to obtain $\bI^{(k)}_{\mathrm{ss}}$ without noise injection. We further compare addition as a noise injection method in Tab. \ref{tab:ablation}.
The soft weight modulation mask for the learning branch is then computed using the same functions as in the pseudo-labeling branch:
\begin{equation}
    \cM^{(k)}_{\mathrm{lrn}} = \s( \cG_1(\bI^{(k)}_{\mathrm{lrn}}) \times \cG_2(\bI^{(k)}_{\mathrm{lrn}})).
\end{equation}
\vspace{-10pt}
\begin{algorithm}[!htp]
\centering
\caption{Domain-guided weight modulation}\label{alg:hyper}
\scriptsize
\begin{algorithmic}[1]
\Require Number of epochs $E$, weak augmentation $\a$, strong augmentation $\cA$, pseudo labeling threshold $\tau$
\For{epochs $1, \ldots, E$}
\For{minibatch indices $(B^{(k)}_\ell, \: B^{(k)}_u)$}
\State \# Compute the domain information vector
\State $\bI^{(k)} \gets \frac{1}{|B^{(k)}_u|} \sum_{i\in B^{(k)}_u} f(\bu^{(k)}_i)$
\State \# Compute the pseudo labeling classifier
\State $\bI^{(k)}_{\mathrm{ss}} \gets \cD(\cE(\bI^{(k)}))$
\State $\cM^{(k)}_{\mathrm{ss}} \gets \s(\cG_1(\bI^{(k)}_{\mathrm{ss}}) \times \cG_2(\bI^{(k)}_{\mathrm{ss}}))$
\State $W^{(k)}_{\mathrm{ss}} \gets W \odot \cM^{(k)}_{\mathrm{ss}}$
\State \# Compute the modulated learning classifier
\State $\bI^{(k)}_{\mathrm{lrn}} \gets \cD(\cE(\bI^{(k)}) + \cN(0, \e^2 I))$
\State $\cM^{(k)}_{\mathrm{lrn}} \gets \s(\cG_1(\bI^{(k)}_{\mathrm{lrn}}) \times \cG_2(\bI^{(k)}_{\mathrm{lrn}}))$
\State $W^{(k)}_{\mathrm{lrn}} \gets W \odot \cM^{(k)}_{\mathrm{lrn}}$
\State \# Pseudolabel the unlabelled data
\State $B^{(k)}_{\mathrm{ss}} \gets \{\}$
\For{$i \in B^{(k)}$}
\If{$\max \mathrm{softmax}(W^{(k)}_{\mathrm{ss}} f(\bu^{(k)}_i)) \geq \tau$}
\State $\tilde{y}^{(k)}_i \gets \argmax W^{(k)}_{\mathrm{ss}} f(\a(\bu^{(k)}_i))$
\State $B^{(k)}_{\mathrm{ss}} \gets B^{(k)}_{\mathrm{ss}} \cup \{i\}$
\EndIf
\EndFor
\State \# Compute the CE loss
\State {\small $\cL_u \gets \frac{1}{|B^{(k)}_{\mathrm{ss}}|}\sum_{i\in B^{(k)}_{\mathrm{ss}}} \mathrm{CE}(W^{(k)}_{\mathrm{lrn}}f(\cA(\bu^{(k)}_i)), \: \tilde{y}^{(k)}_i)$}
\State {\small $\cL_\ell \gets \frac{1}{|B^{(k)}_\ell|} \sum_{i\in B^{(k)}_\ell} \mathrm{CE}(W^{(k)}_{\mathrm{lrn}} f(\a(\bx^{(k)}_i)), y^{(k)}_i)$}
\State \# Update the learned components
\State $\mathrm{update}(f, W, \cE, \cD, \cG_1, \cG_2; \: \cL_\ell + \cL_u)$
\EndFor
\EndFor
\State \Return Trained components $f$, $W$, $\cE$, $\cD$, $\cG_1$, $\cG_2$
\end{algorithmic}

\end{algorithm}

The noise introduced in the modulation mask generation due to the lower-rank structure and the perturbed domain information vector can be seen as a form of consistency regularization \cite{tarvainen2017mean}.
Further, we empirically show that this noise-injected encoder/decoder-like structure outperformed learning a general map of $\bI^{(k)}\mapsto \cM^{(k)}_{\mathrm{ss}}$ (Table~\ref{tab:ablation}). 

\noindent\textbf{Loss computation \& model update:}
Again following FixMatch \cite{sohn2020fixmatch}, we make training predictions on \emph{strongly} augmented versions of the unlabelled data, and optimize so that these predictions match the pseudolabels. We accomplish this with the cross-entropy loss, averaged over the pseudo labeled points $B^{(k)}_{\mathrm{ss}}$. Thus, loss for the unlabelled points is: 
\begin{equation}
    \cL_u = \frac{1}{|B^{(k)}_{\mathrm{ss}}|} \sum_{i \in B^{(k)}_{\mathrm{ss}}} \mathrm{CE}(W^{(k)}_{\mathrm{lrn}} f(\cA(\bu^{(k)}_i)), \: \tilde{y}^{(k)}_i). 
\end{equation}
Here, $\cA$ is strong augmentation function \cite{sohn2020fixmatch}. For notational convenience, we define the cross entropy loss $\mathrm{CE}: \R^C \times \cY \rightarrow \R$ so that the first argument is the model logits and the second argument is the target label.
We also use the labeled examples to compute the standard cross-entropy loss. The labelled examples only use the weak augmentation function $\a$ \cite{sohn2020fixmatch}, so the loss for the labelled points is
\begin{equation}
    \cL_\ell = \frac{1}{|B^{(k)}_\ell|} \sum_{i\in B^{(k)}_\ell} \mathrm{CE}(W^{(k)}_{\mathrm{lrn}} f(\a(\bx^{(k)}_i)), \: y^{(k)}_i). 
\end{equation}

Treating $\tilde{y}^{(k)}_i$ as constants, we can backpropagate through the loss $\cL_u + \cL_\ell$ and update the parameters for the learned components $f$, $W$, $\cE$, $\cD$, $\cG_1$, and $\cG_2$ using standard optimization procedures such as SGD or Adam. We denote such a generic update of the learned components based on the loss as $\mathrm{update}(f, W, \cE, \cD, \cG_1, \cG_2; \: \cL_u + \cL_\ell)$.

\noindent\textbf{Inference at Test Time:}
At test time, we make inference using the \emph{domain-shared} (unmodulated) classifier weights $W$, using the unaugmented input. That is, given a test point $\bx^\cT$ from the target domain, our prediction is $\argmax W f(\bx^\cT)$. 

\subsection{Effects of Weight Modulation}
\label{subsection:Effects of Weight Modulation}
Let $v_{\mathrm{cls}} = \cG_1(\bI)$ and $v_f = \cG_2(\bI)$. (Here, $\bI$ stands in for either $\bI^{(k)}_{\mathrm{ss}}$ or $\bI^{(k)}_{\mathrm{lrn}}$.) Observe that $v_f$ partitions the learned features into two complementary subsets: those features $J_+ = \{j \in [d] \: : \: v_f[j] \geq 0\}$, and those $J_- = \{j \in [d] \: : \: v_f[j] < 0\}$. For each class $c \in \cY$, if $v_{\mathrm{cls}}[c] > 0$ increases, the features in $J_+$ will be up-weighted relative to the other features by the modulation (assigned weight $\geq1/2$, closer to $1$) while the features in $J_-$ will be down-weighted relative to the other features (assigned weight $<1/2$, closer to $0$). Conversely, if $v_{\mathrm{cls}}[c] < 0$ decreases, the $J_-$ features will be up-weighted and $J_+$ will be down-weighted. Since $v_f$ depends on the domain information vector, these complementary sets of features are domain-specific. The rate at which the reweighting occurs depends on the magnitude of each $v_f[j]$.
The up- and down-weighting of different subsets of the features have two effects. The first concerns pseudo-label generation and prediction: if $v_{\mathrm{cls}}[c] > 0$ and the features in $J_+$ are up-weighted relative to the other features for class $c$, this means that the $J_+$ features will be more heavily relied up to predict class $c$, and vice-versa. This allows the domain-shared classifier weights $W$ to adapt to particulars of each domain when generating  pseudo-labels.

The second effect reinforces this increased reliance on certain features during learning.
Let $z_{\mathrm{mod}} = (W \odot \cM) f(\bx)$ be the model logits on input $\bx$ when weight modulation is applied, and $z_{\mathrm{std}} = W f(\bx)$ be the model logits on $\bx$ without applying weight modulation. The resulting loss gradients w.r.t the domain-shared classifier weights $W$ are:
\begin{align*}
    \nb_W [ \mathrm{CE}(z_{\mathrm{mod}}, \: y) ] &= (\nb_z \mathrm{CE}(z_{\mathrm{mod}}, \: y) \times f(\bx)^\T) \odot \cM, \\
    \nb_W [ \mathrm{CE}( z_{\mathrm{std}}, \: y )] &= \nb_z \mathrm{CE}(z_{\mathrm{std}}, \: y) \times f(\bx)^\T,
\end{align*}
where $\nb_z\mathrm{CE}( z, y ) \in \R^C$ is the gradient of the CE loss with respect to the logits $z$. Observe that the weight modulation is applied also to the gradient update for the domain-shared classifier weights, reinforcing the classifier's reliance on the (relatively) up weighted features.

\section{Experiments}
\label{secttion:Experiments}

\noindent\textbf{Datasets:}
We conducted experiments on six widely used DG datasets: PACS \cite{li2017deeper}, OfficeHome \cite{venkateswara2017deep}, VLCS \cite{fang2013unbiased}, DigitsDG \cite{zhou2020deep}, TerraIncognita \cite{beery2018recognition} and DoamainNet \cite{peng2019moment}. See suppl.  for a detailed description of datasets.

\begin{table*}[]

  \resizebox{\textwidth}{!}{
  \setlength{\tabcolsep}{0.9mm}{
  \begin{tabular}{c|cccccc|cccccc}
    \toprule
    \multirow{2}{*}{\textbf{Method}}   &   \multicolumn{6}{c|}{\textbf{5 labels}}  &   \multicolumn{6}{c}{\textbf{10 labels}}   \\
    ~   &   \textbf{PACS}  &   \textbf{OfficeHome}    &   \textbf{VLCS}    &   \textbf{DigitsDG}  &   \textbf{TerraInc}  &   \textbf{DomainNet}   &   \textbf{PACS}  &   \textbf{OfficeHome}    &   \textbf{VLCS}    &   \textbf{DigitsDG}  &   \textbf{TerraInc}  &   \textbf{DomainNet}   \\ \midrule
    ERM &   51.2$_{\pm3.0}$    &   51.7$_{\pm0.6}$    &   67.2$_{\pm1.8}$    &   22.7$_{\pm1.0}$    &   22.9$_{\pm3.0}$    &   23.5$_{\pm0.2}$  &  59.8$_{\pm2.3}$   &   56.7$_{\pm0.8}$    &   68.0$_{\pm0.3}$    &   29.1$_{\pm2.9}$    &   23.5$_{\pm1.2}$    &   29.4$_{\pm0.1}$ \\
    EntMin &   55.9$_{\pm2.1}$    &   52.7$_{\pm0.6}$    &   66.5$_{\pm1.0}$    &   28.7$_{\pm1.3}$    &   21.4$_{\pm3.5}$    &   24.1$_{\pm0.3}$ &   64.0$_{\pm2.2}$    &   57.0$_{\pm0.8}$    &   66.2$_{\pm0.2}$    &   39.3$_{\pm2.8}$    &   26.6$_{\pm2.6}$   &   28.5$_{\pm0.1}$ \\
    MeanTeacher &   55.3$_{\pm4.0}$    &   50.9$_{\pm0.7}$    &   66.4$_{\pm1.0}$    &   28.5$_{\pm1.4}$    &   20.9$_{\pm2.5}$    &   24.2$_{\pm0.2}$    &   61.5$_{\pm1.4}$    &   55.9$_{\pm0.5}$    &   66.2$_{\pm0.4}$    &   38.8$_{\pm2.9}$    &   25.0$_{\pm2.8}$    &   28.6$_{\pm0.1}$\\
    FixMatch &   73.4$_{\pm1.3}$    &   55.1$_{\pm0.5}$    &   69.9$_{\pm0.6}$    &   56.0$_{\pm2.2}$    &   28.9$_{\pm2.3}$    &   26.7$_{\pm0.2}$   &   76.6$_{\pm1.2}$    &   57.8$_{\pm0.3}$    &   70.0$_{\pm2.1}$    &   66.4$_{\pm1.4}$   &   30.5$_{\pm1.2}$ &   29.2$_{\pm0.5}$   \\
    FBCSA & 77.3$_{\pm1.1}$ & 55.8$_{\pm0.2}$ & 71.3$_{\pm0.7}$ & 62.0$_{\pm1.5}$ & 33.2$_{\pm2.0}$ & - & 78.2$_{\pm1.2}$ & 59.0$_{\pm0.4}$ & 72.2$_{\pm1.0}$ & 70.4$_{\pm1.4}$ & 34.7$_{\pm1.9}$ & - \\
    StyleMatch &   78.4$_{\pm1.1}$   &   56.3$_{\pm0.3}$    &   72.5$_{\pm1.5}$    &   55.7$_{\pm1.6}$    &   28.7$_{\pm2.7}$    &   25.5$_{\pm0.1}$ &   79.4$_{\pm0.9}$    &   59.7$_{\pm0.2}$    &   73.3$_{\pm0.6}$    &   64.8$_{\pm1.9}$    &   29.9$_{\pm2.8}$    &   29.1$_{\pm0.4}$\\
    \midrule
    EntMin+Ours &   57.7$_{\pm3.0}$    &   54.3$_{\pm0.6}$    &   67.0$_{\pm0.9}$    &   31.1$_{\pm2.2}$    &   23.6$_{\pm2.8}$    &   25.6$_{\pm0.2}$    &   63.9$_{\pm1.3}$    &   58.2$_{\pm0.3}$    &   66.5$_{\pm0.2}$    &   42.2$_{\pm2.3}$    &   28.2$_{\pm0.7}$    &   29.7$_{\pm0.2}$\\
    MeanTeacher+Ours &   55.9$_{\pm2.9}$    &   53.2$_{\pm0.8}$    &   66.0$_{\pm1.0}$    &   31.5$_{\pm2.1}$    &   22.3$_{\pm2.3}$  & 25.7$_{\pm0.2}$   &   62.3$_{\pm1.0}$    &   57.6$_{\pm0.4}$    &   66.5$_{\pm0.4}$    &   42.8$_{\pm1.1}$    &   28.1$_{\pm0.9}$    &   \textbf{29.9$_{\pm0.2}$}\\
    FixMatch+Ours &   77.9$_{\pm0.8}$    & 56.2$_{\pm0.2}$    &   \textbf{75.2$_{\pm0.9}$}    &   57.4$_{\pm1.5}$   &   31.0$_{\pm2.8}$    &   \textbf{26.9$_{\pm0.2}$}  &   78.4$_{\pm1.0}$    &   59.7$_{\pm0.3}$    &   75.2$_{\pm0.7}$    &   68.4$_{\pm1.5}$   &   32.1$_{\pm2.4}$   &   29.6$_{\pm0.2}$\\
    FBCSA+Ours & 77.9$_{\pm0.9}$ &56.2$_{\pm0.2}$ &71.8$_{\pm1.1}$&\textbf{63.3$_{\pm1.6}$}&\textbf{33.8$_{\pm1.4}$}&-&78.9$_{\pm0.8}$&59.7$_{\pm0.3}$&\textbf{75.5$_{\pm0.5}$}&\textbf{71.3$_{\pm1.3}$}& \textbf{35.0$_{\pm1.7}$} & - \\
    
    StyleMatch+Ours &   \textbf{79.4$_{\pm0.6}$}    &   \textbf{56.8$_{\pm0.3}$}    &   73.5$_{\pm0.4}$   &   56.6$_{\pm0.6}$    &   30.0$_{\pm3.3}$    &   26.7$_{\pm0.3}$  &   \textbf{80.7$_{\pm0.8}$}    &   \textbf{60.0$_{\pm0.1}$}    &   74.1$_{\pm0.8}$   &   66.3$_{\pm1.1}$    &   30.1$_{\pm2.8}$    &   \textbf{29.9$_{\pm0.2}$}\\
    \bottomrule
    \end{tabular}
    }
}    
  \caption{Comparison with the SOTA SSL-based SSDG baselines and SSDG methods under the first setting.
  When averaged across datasets we achieve a performance gain of $+\textbf{2.4}\%$ and $+\textbf{2.1}\%$ in 5,10 labels per class setting over the baseline FixMatch. } 
  \label{tab:setting1}
    \end{table*}

\noindent\textbf{Training, implementation details \& evaluation protocol:} We follow the same training settings as in StyleMatch \cite{zhou2023semi}. ImageNet \cite{deng2009imagenet} pretrained ResNet-18 \cite{he2016deep} is used as the backbone for all the experiments. SGD is used as the optimizer for both the backbone and the classifier with the initial learning rates of 0.003 and 0.01, respectively. Both learning rates are decayed using cosine annealing. We train all methods on all datasets for 20 epochs except for TerraIncognita and DomainNet. For TerraIncognita and DomainNet we train for 10 epochs.  16 labeled data and 16 unlabeled data from each source domain are randomly sampled to construct a minibatch. The supervised loss is computed using the labelled subset of the minibatch, and the complete minibatch (without ground truth labels in the labeled subset) is used to compute the unsupervised loss \cite{zhou2023semi}. For a fair comparison, we chose methods that share similar SSDG settings as ours and their code is publicly available. We report top-1 accuracy over 5 independent trials. We provide ablations and all other experiments (unless otherwise specified) under 10 labels setting on OfficeHome Dataset. We adopt leave-one-domain-out evaluation protocol to report results as it is widely used in DG \cite{Gulrajani2021InSO} and SSDG \cite{zhou2023semi, galappaththige2024towards}. 

\noindent\textbf{First setting:}  
Here, each source domain has only a few labeled examples (either 5 or 10 labels per class) and the rest of the examples are unlabeled (Tab.~\ref{tab:setting1}). Under both 5 and 10 label scenarios Our method consistently provides notable gains over the baselines. For instance, in OfficeHome dataset (10 labels), our method delivers an absolute gain of 1.9\% over FixMatch. When available data are further constrained, e.g., in the VLCS dataset (5 labels), our method provides a significant gain of 5.3\% over the FixMatch. In summary, by integrating our method, the performances of strong baselines of SSL and SSDG methods under both 10 and 5 label scenarios can be further boosted, validating its effectiveness and versatility. Note that we are unable to produce FBCSA \cite{galappaththige2024towards} results on DomainNet as GPU memory runs out due to the need to create domain-aware class prototypes for 345 classes in all 5 source domains.

\noindent\textbf{Second setting:} 
Here, only the data from one source domain is fully labeled and the data from others are completely unlabeled. We show the comparison with other methods in Tab.~\ref{tab:one_fully_labeled}. Note that FBCSA\cite{galappaththige2024towards} cannot be used under this setting as it needs labeled data points from all source domains to create domain-aware class prototypes. Our method boosts the performance of existing baselines in all instances. 
Overall, in this more challenging setting, our method's gains are even higher than the first setting.

\begin{table}[]
  \centering
  \scalebox{0.66}{
  \setlength{\tabcolsep}{.5mm}{
  \begin{tabular}{c|cccccc}
    \toprule
    \textbf{Method}  &   \textbf{PACS}  &   \textbf{OfficeHome}    &   \textbf{VLCS}    &   \textbf{Digits}  &   \textbf{TerraInc}  &   \textbf{DomainNet}    \\\midrule
    ERM &   69.8$_{\pm1.8}$   &   61.7$_{\pm0.4}$    &   60.8$_{\pm0.7}$    &   36.7$_{\pm0.7}$   &   40.0$_{\pm2.3}$    &   33.1$_{\pm0.1}$\\
    EntMin &   76.9$_{\pm1.8}$    &   61.9$_{\pm0.2}$   &   55.6$_{\pm0.7}$    &   40.1$_{\pm1.0}$    &   39.1$_{\pm2.7}$    &   35.2$_{\pm0.1}$\\
    MeanTeacher &   74.6$_{\pm1.4}$   &   60.4$_{\pm0.2}$   &   55.9$_{\pm0.4}$    &   38.8$_{\pm0.7}$   &   38.3$_{\pm1.4}$   &   36.8$_{\pm0.1}$\\
    
    FixMatch &   79.9$_{\pm1.4}$   &   62.1$_{\pm0.2}$    &   58.9$_{\pm1.3}$    &   53.4$_{\pm0.9}$    &   39.7$_{\pm3.3}$    &   32.2$_{\pm0.3}$\\
    StyleMatch &   80.8$_{\pm3.3}$   &   63.3$_{\pm0.4}$   &   63.3$_{\pm2.3}$    &   49.3$_{\pm0.3}$   &   34.1$_{\pm3.0}$   &   30.8$_{\pm0.1}$\\
    \midrule
    EntMin+Ours &   76.7$_{\pm1.3}$    &   64.0$_{\pm0.2}$   &   55.7$_{\pm0.8}$   &   40.5$_{\pm1.0}$   &   42.1$_{\pm1.3}$   &   36.5$_{\pm0.2}$\\
    MeanTeacher+Ours &   75.0$_{\pm1.6}$   &   63.1$_{\pm0.1}$   &   55.9$_{\pm1.2}$    &   39.2$_{\pm0.7}$  &   40.5$_{\pm1.0}$    &   \textbf{38.1$_{\pm0.1}$}\\
    FixMatch+Ours &   82.1$_{\pm0.9}$  &   \textbf{64.2$_{\pm0.2}$}    &   \textbf{65.7$_{\pm1.8}$}    &   \textbf{55.6$_{\pm0.9}$}    &   \textbf{43.3$_{\pm1.1}$}    &   32.6$_{\pm0.1}$\\
    StyleMatch+Ours &   \textbf{83.8$_{\pm0.5}$}    &   63.5$_{\pm0.3}$    &   63.9$_{\pm2.9}$    &   49.5$_{\pm0.8}$   &   36.2$_{\pm1.0}$   &   30.7$_{\pm0.1}$\\
    \bottomrule
    \end{tabular} }
    }
  \caption{Comparison with SOTA SSL-based SSDG baselines and SSDG methods under the second setting. When averaged across datasets we achieve a  gain of +\textbf{3.1}\% over the baseline FixMatch.}
  \label{tab:one_fully_labeled}
  \vspace{-1em}
\end{table}

\begin{table*}[t]

    \centering
    \scalebox{1}{
    \resizebox{\textwidth}{!}{
    \begin{tabular}{cccccccccc}
    \toprule
         \multirow{2}{*}{\textbf{Domain Shift}}  &   \multirow{2}{*}{\textbf{Dataset}}   &  \multicolumn{8}{c}{\textbf{Method}}   \\ \cmidrule{3-10}
         ~  &   ~    & EntM.  & MeanT.  & FixM.  & StyleM.  & EntM.+\textbf{Ours}  & MeanT.+\textbf{Ours}    & FixM.+\textbf{Ours}    & StyleM.+\textbf{Ours}    \\ \midrule
         Style Shifts   &   OfficeHome,PACS    & 60.5  & 58.7  & 67.2  & 69.5  & 61.1  & 60.0  & 69.0  &  \textbf{70.35} \\ 
         Background Shifts   &   VLCS,Digits    & 52.7  & 52.5  & 68.2  & 69.1  & 54.4  & 54.7  & \textbf{71.5}  &  70.2 \\

         Corruption Shift   &   TerraInc    & 26.6 & 25.0  & 30.5  & 29.9 & 28.2  & 28.1 & \textbf{31.9}  &  30.1\\
    \bottomrule
    \end{tabular}
    }}

   \caption{Performance under different types of distribution shifts.}
    \label{tab:various_dis_shift} \vspace{-1em}
\end{table*}

\noindent\textbf{Contribution of different components:} We conduct a comprehensive ablation on the key components of our method (see Tab.~\ref{tab:ablation}) under 10 Labels settings. 
We show that employing separate classifiers for each domain does not improve the SSDG performance. This is likely because the number of available labeled data points for each classifier becomes further constrained and there is a greater chance of overfitting, especially in 5-label settings. 
Then, we show the importance of our noise-injected encoder-decoder (NIED) and low-rank (LR) decomposed structure. 
Learning a general map, using a single MLP, from domain information to soft mask $\bI^{(k)}\mapsto \cM^{(k)}_{ss}$ improves over separate classifiers, but it is inferior to employing our noise-injected encoder-decoder (NIED). Also, NIED without injecting noise reveals deteriorated performance. We further compare noise addition against noise concatenation. Finally, our proposal of coupling NIED with the LR decomposed structure shows the best performance. 

\begin{table}[]

\centering
\scalebox{0.8}{
\begin{tabular}{lc}
\toprule
\textbf{Method   }                                                                                   & \textbf{Average} \\ \midrule
Baseline (FixMatch \cite{sohn2020fixmatch})      & 57.8   \\
Baseline + Separate domain classifiers & 57.7  \\
Baseline + General map ($\bI^{(k)}\mapsto \cM^{(k)}_{ss}$) &  58.5 \\
Baseline + NIED (without noise) & 58.4 \\
Baseline + NIED (noise addition) & 58.6 \\
Baseline + NIED (noise concatenation) & \underline{58.8} \\

Baseline + NIED + LR (Ours) & \textbf{59.7} \\
   \bottomrule   

\end{tabular}}
\caption{Contribution of key components.}

\label{tab:ablation}
\end{table}

\noindent\textbf{On aggregating domain-level information:} We compare various approaches for aggregating the domain information for each domain in a minibatch (see Tab. ~\ref{tab:domain_analysis}). 1) Train a separate backbone (ResNet-18) as an auxiliary branch to predict the domain label for an image and then compute the mean of the features from the auxiliary backbone. 2) Train a domain projection head (2-layer MLP), after the baseline's backbone, to predict the domain label for an image and compute the mean of representation from the domain projection head. 3) Utilize only the principal eigenvector of the variance-covariance matrix formed from the features produced by the baseline's backbone for a given domain, 4) Use the mean over all the eigenvectors of the variance-covariance matrix of the features produced by the backbone for a given domain, 
5) Finally, we simply compute the mean of the features produced by the baseline's backbone for a given domain (Eq.~\ref{eq:domain-info}). 
We observe that a simple central tendency measure i.e. the mean over backbone features for aggregating domain-level information provides competitive performances. Therefore, we stick to computing the mean in our method, thereby avoiding further computations or additional learnable parameters.

\begin{table}[]

\centering
\scalebox{0.8}{
\begin{tabular}{lc}
\toprule
\textbf{Method   }                                                                                   & \textbf{Average} \\ \midrule
Auxiliary backbone & 56.6   \\
Auxiliary projection head & \textbf{59.8}  \\
Principal eigenvector  &  59.2 \\
Mean eigenvectors & 58.8 \\
central tendency (Ours) & \underline{59.7}\\

   \bottomrule   

\end{tabular}}
\caption{Approaches for domain information aggregation.}
\label{tab:domain_analysis}
\end{table}

\noindent\textbf{Performance with different backbones:} Tab.~\ref{tab:other_backbones} reports results with several stronger backbones. Our method \emph{consistently improves} over baseline even with stronger backbones. 

\begin{table}[h]
    \centering
    \tabcolsep=0.05cm

    \scalebox{0.75}{
        \begin{tabular}{lcccccc}
            \toprule
            \textbf{Algorithm}  & \textbf{RN18} & \textbf{RN50} & \textbf{RN101} & \textbf{Vit-S/32} & \textbf{Vit-B/32} & \textbf{CLIP-B/32} \\
            \midrule
            FixMatch\cite{sohn2020fixmatch}                & 57.8$_{\pm0.3}$       & 61.3$_{\pm0.4}$ & 62.8$_{\pm0.2}$ & 63.7$_{\pm0.5}$ & 72.0$_{\pm0.4}$ & 75.3$_{\pm0.6}$ \\
            FixM. +Ours                & \textbf{59.7}$_{\pm0.3}$  & \textbf{64.2$_{\pm0.2}$} & \textbf{66.7$_{\pm0.2}$} & \textbf{65.4$_{\pm0.3}$} & \textbf{75.0$_{\pm0.3}$}& \textbf{78.6$_{\pm0.1}$} \\
            \bottomrule
        \end{tabular}
    
    }

\caption{Results with different backbones.}
    \label{tab:other_backbones} \vspace{-1em}
\end{table}

\noindent\textbf{Performance with varing number of labels:}  We report how the performance scale when we increase the number of labeled data points per-class on OfficeHome dataset (Tab.~\ref{tab:different_labels_per_class}). Our method \emph{consistently improves} over its baseline\cite{sohn2020fixmatch} in all the per-class labels settings. 

\begin{table}[h]
    \centering
    \tabcolsep=0.1cm
    \scalebox{0.8}{
        \begin{tabular}{lccccc}
            \toprule
            \textbf{Algorithm}  & \textbf{5} & \textbf{10} & \textbf{15} & \textbf{20} & \textbf{25} \\
            \midrule
            FixMatch\cite{sohn2020fixmatch}                & 55.1$_{\pm0.5}$       & 57.8$_{\pm0.3}$       & 59.2$_{\pm0.2}$ & 59.9$_{\pm0.4}$ & 60.2$_{\pm0.4}$ \\

            FixM. +Ours                & \textbf{56.5}$_{\pm0.3}$   & \textbf{59.7}$_{\pm0.3}$ & \textbf{61.1}$_{\pm0.4}$ & \textbf{62.0}$_{\pm0.2}$ & \textbf{62.4}$_{\pm0.1}$ \\
            \bottomrule
        \end{tabular}

    }
\caption{Results with different numbers of labels per-class settings}

    \label{tab:different_labels_per_class} \vspace{-1em}

\end{table}

\noindent\textbf{Training efficiency:} We compare our method with existing SSDG methods \cite{zhou2023semi, galappaththige2024towards} in terms of training efficiency. Unlike StyleMatch which adds 203.3\% additional overhead over its baseline FixMatch mostly due to its style transferring module in multi-view consistency branch \cite{zhou2023semi}, we only add a small overhead as little as 13.33\%. We report the average time per epoch in seconds on a single A6000 GPU for the OfficeHome dataset in Tab.~\ref{tab:training-eff} for 10 labels setting. 

\begin{table}[h]

\centering
\scalebox{0.74}{

\begin{tabular}{lcc}
\toprule
\textbf{Method}                                                                                      & \textbf{Average time/epoch} & \textbf{Overhead} \\ \midrule
FixMatch \cite{sohn2020fixmatch} & 22.5 & - \\
FBCSA \cite{galappaththige2024towards} &36.5 & 58.22\% \\
StyleMatch\cite{zhou2023semi} &68.25& 203.33\% \\
FixMatch + Ours & 25.5 & \textbf{13.33\%} \\

\bottomrule   

\end{tabular}
}
\caption{Training overhead over the baseline FixMatch.}
\label{tab:training-eff} \vspace{-1em}

\end{table}

\noindent\textbf{Performance under various distribution shifts:} 
Tab. ~\ref{tab:various_dis_shift} compares the performance under different distribution shifts e.g., style shifts, background shifts, and corruption shifts. 
Existing SSDG methods (StyleMatch \cite{zhou2023semi}) assume some style distribution shifts in the source domain and hence struggle under corruption or background shifts. Unlike StyleMatch, our approach shows significant gains over baselines under all distribution shifts.  

\begin{figure}[h]
  \centering   \includegraphics[width=.9\linewidth]{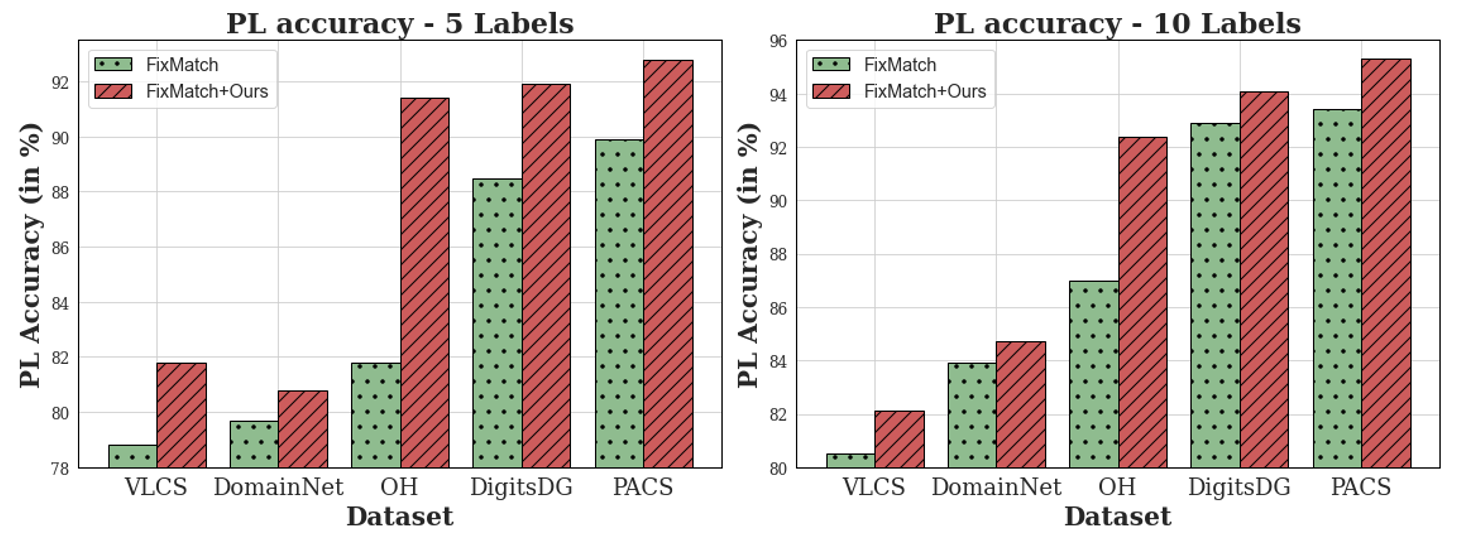}
   \caption{PL accuracy during training for baseline \cite{sohn2020fixmatch} and ours.}  \vspace{-1em} \label{fig:PL_acc_overall_all_datasets}
\end{figure}

\begin{figure}[h]
  \centering
   \includegraphics[width=.75\linewidth]{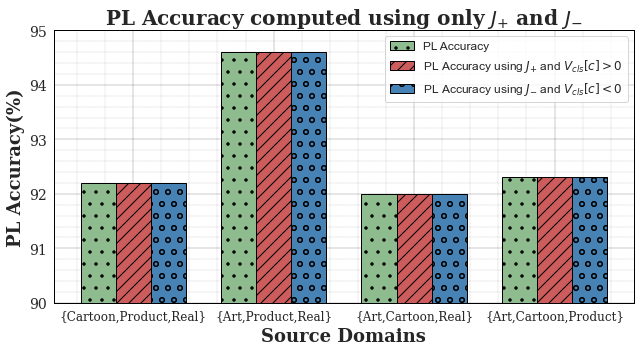}
   \caption{PL accuracy when computed using using features in $J_+$ for classes $c$ with $v_{\mathrm{cls}}[c] > 0$ and features in $J_-$ for classes $c$ with $v_{\mathrm{cls}}[c] < 0$ on OH dataset (10 labels). }
     
     \label{fig:verify_1} \vspace{-1em}
\end{figure}

\noindent\textbf{Improved PL accuracy: } We plot the pseudo-labeling accuracy after the thresholding process \cite{sohn2020fixmatch} in Fig.~\ref{fig:PL_acc_overall_all_datasets} on both 5 and 10 label settings. Our proposed method can improve the pseudo-labeling in different datasets which exhibit various distribution shifts. The reason is that the weight modulation in our method tends to reduce the model's maximum confidence when computing pseudo-labels. The result is that only highly accurate pseudo-labels will make it past the confidence threshold.
\noindent\textbf{Verification of \S\ref{subsection:Effects of Weight Modulation}:} We empirically verify our claim on the effect of weight modulation on pseudo-labeling (PL). We compute the PLs where the logits are computed just using features in $J_+$ for classes $c$ with $v_{\mathrm{cls}}[c] > 0$ and just using features in $J_-$ for classes $c$ with $v_{\mathrm{cls}}[c] < 0$, while comparing these to the pseudo-labels actually generated by the method. 
The two sets of pseudo-labels are very similar (Fig.~\ref{fig:verify_1}), indicating that the mask causes the pseudo-labeler to rely on specific subsets of features depending on the domain and class, as claimed.

\section{Conclusion and Limitations}

Towards tackling a relatively understudied problem of semi-supervised domain generalization, we proposed a domain-guided weight modulation method that learns a soft modulation mask for imparting domain-level information into the classifier on-the-fly during training. Thorough experiments on six challenging and diverse benchmarks showcase the superlative performance of our method over strong SSL-based SSDG baselines. A potential limitation of our method is that it would require pertinent modifications to be applicable to single-source semi-supervised DG, which is left for future work.

{\small
\bibliographystyle{ieee_fullname}
\bibliography{egbib}

\begin{thebibliography}{10}\itemsep=-1pt

\bibitem{abuduweili2021adaptive}
Abulikemu Abuduweili, Xingjian Li, Humphrey Shi, Cheng-Zhong Xu, and Dejing Dou.
\newblock Adaptive consistency regularization for semi-supervised transfer learning.
\newblock In {\em Proceedings of the IEEE/CVF Conference on Computer Vision and Pattern Recognition}, pages 6923--6932, 2021.

\bibitem{Agarap2018DeepLU}
Abien~Fred Agarap.
\newblock Deep learning using rectified linear units (relu).
\newblock {\em ArXiv}, abs/1803.08375, 2018.

\bibitem{NIPS2014_66be31e4}
Philip Bachman, Ouais Alsharif, and Doina Precup.
\newblock Learning with pseudo-ensembles.
\newblock In Z. Ghahramani, M. Welling, C. Cortes, N. Lawrence, and K.Q. Weinberger, editors, {\em Advances in Neural Information Processing Systems}, volume~27. Curran Associates, Inc., 2014.

\bibitem{meta_learning2}
Yogesh Balaji, Swami Sankaranarayanan, and Rama Chellappa.
\newblock Metareg: Towards domain generalization using meta-regularization.
\newblock {\em Advances in neural information processing systems}, 31, 2018.

\bibitem{beery2018recognition}
Sara Beery, Grant Van~Horn, and Pietro Perona.
\newblock Recognition in terra incognita.
\newblock In {\em European Conference on Computer Vision}, pages 456--473, 2018.

\bibitem{berthelot2019mixmatch}
David Berthelot, Nicholas Carlini, Ian Goodfellow, Nicolas Papernot, Avital Oliver, and Colin~A Raffel.
\newblock Mixmatch: A holistic approach to semi-supervised learning.
\newblock {\em Advances in neural information processing systems}, 32, 2019.

\bibitem{blanchard2011generalizing}
Gilles Blanchard, Gyemin Lee, and Clayton Scott.
\newblock Generalizing from several related classification tasks to a new unlabeled sample.
\newblock {\em NeurIPS}, 24:2178--2186, 2011.

\bibitem{carlucci2019domain}
Fabio~M Carlucci, Antonio D'Innocente, Silvia Bucci, Barbara Caputo, and Tatiana Tommasi.
\newblock Domain generalization by solving jigsaw puzzles.
\newblock In {\em CVPR}, pages 2229--2238, 2019.

\bibitem{chen2023softmatch}
Hao Chen, Ran Tao, Yue Fan, Yidong Wang, Jindong Wang, Bernt Schiele, Xing Xie, Bhiksha Raj, and Marios Savvides.
\newblock Softmatch: Addressing the quantity-quality tradeoff in semi-supervised learning.
\newblock In {\em The Eleventh International Conference on Learning Representations}, 2023.

\bibitem{deng2009imagenet}
Jia Deng, Wei Dong, Richard Socher, Li-Jia Li, Kai Li, and Li Fei-Fei.
\newblock Imagenet: A large-scale hierarchical image database.
\newblock In {\em 2009 IEEE conference on computer vision and pattern recognition}, pages 248--255. Ieee, 2009.

\bibitem{dosovitskiy2020image}
Alexey Dosovitskiy, Lucas Beyer, Alexander Kolesnikov, Dirk Weissenborn, Xiaohua Zhai, Thomas Unterthiner, Mostafa Dehghani, Matthias Minderer, Georg Heigold, Sylvain Gelly, et~al.
\newblock An image is worth 16x16 words: Transformers for image recognition at scale.
\newblock {\em arXiv preprint arXiv:2010.11929}, 2020.

\bibitem{meta_learning3}
Qi Dou, Daniel Coelho~de Castro, Konstantinos Kamnitsas, and Ben Glocker.
\newblock Domain generalization via model-agnostic learning of semantic features.
\newblock {\em Advances in neural information processing systems}, 32, 2019.

\bibitem{dumoulin2017a}
Vincent Dumoulin, Jonathon Shlens, and Manjunath Kudlur.
\newblock A learned representation for artistic style.
\newblock In {\em International Conference on Learning Representations}, 2017.

\bibitem{fang2013unbiased}
Chen Fang, Ye Xu, and Daniel~N Rockmore.
\newblock Unbiased metric learning: On the utilization of multiple datasets and web images for softening bias.
\newblock In {\em ICCV}, pages 1657--1664, 2013.

\bibitem{dropout}
Yarin Gal and Zoubin Ghahramani.
\newblock Dropout as a bayesian approximation: Representing model uncertainty in deep learning.
\newblock In {\em Proceedings of the 33rd International Conference on International Conference on Machine Learning - Volume 48}, ICML'16, page 1050–1059. JMLR.org, 2016.

\bibitem{galappaththige2024towards}
Chamuditha~Jayanga Galappaththige, Sanoojan Baliah, Malitha Gunawardhana, and Muhammad~Haris Khan.
\newblock Towards generalizing to unseen domains with few labels.
\newblock In {\em Proceedings of the IEEE/CVF Conference on Computer Vision and Pattern Recognition}, pages 23691--23700, 2024.

\bibitem{grandvalet2004semi}
Yves Grandvalet and Yoshua Bengio.
\newblock Semi-supervised learning by entropy minimization.
\newblock {\em Advances in neural information processing systems}, 17, 2004.

\bibitem{Gulrajani2021InSO}
Ishaan Gulrajani and David Lopez-Paz.
\newblock In search of lost domain generalization.
\newblock {\em ArXiv}, abs/2007.01434, 2021.

\bibitem{he2016deep}
Kaiming He, Xiangyu Zhang, Shaoqing Ren, and Jian Sun.
\newblock Deep residual learning for image recognition.
\newblock In {\em Proceedings of the IEEE conference on computer vision and pattern recognition}, pages 770--778, 2016.

\bibitem{hoffman2018cycada}
Judy Hoffman, Eric Tzeng, Taesung Park, Jun-Yan Zhu, Phillip Isola, Kate Saenko, Alexei Efros, and Trevor Darrell.
\newblock Cycada: Cycle-consistent adversarial domain adaptation.
\newblock In {\em International conference on machine learning}, pages 1989--1998. Pmlr, 2018.

\bibitem{Huang2017ArbitraryST}
Xun Huang and Serge~J. Belongie.
\newblock Arbitrary style transfer in real-time with adaptive instance normalization.
\newblock {\em 2017 IEEE International Conference on Computer Vision (ICCV)}, pages 1510--1519, 2017.

\bibitem{khan2021mode}
Muhammad~Haris Khan, Talha Zaidi, Salman Khan, and Fahad~Shehbaz Khan.
\newblock Mode-guided feature augmentation for domain generalization.
\newblock In {\em Proc. Brit. Mach. Vis. Conf.}, 2021.

\bibitem{laine2017temporal}
Samuli Laine and Timo Aila.
\newblock Temporal ensembling for semi-supervised learning.
\newblock In {\em International Conference on Learning Representations}, 2017.

\bibitem{lee2013pseudo}
Dong-Hyun Lee et~al.
\newblock Pseudo-label: The simple and efficient semi-supervised learning method for deep neural networks.
\newblock In {\em Workshop on challenges in representation learning, ICML}, volume~3, page 896, 2013.

\bibitem{li2017deeper}
Da Li, Yongxin Yang, Yi-Zhe Song, and Timothy~M Hospedales.
\newblock Deeper, broader and artier domain generalization.
\newblock In {\em ICCV}, pages 5542--5550, 2017.

\bibitem{li2019episodic}
Da Li, Jianshu Zhang, Yongxin Yang, Cong Liu, Yi-Zhe Song, and Timothy~M Hospedales.
\newblock Episodic training for domain generalization.
\newblock {\em In ICCV}, 2019.

\bibitem{domain_alignment3}
Haoliang Li, Sinno~Jialin Pan, Shiqi Wang, and Alex~C Kot.
\newblock Domain generalization with adversarial feature learning.
\newblock In {\em Proceedings of the IEEE conference on computer vision and pattern recognition}, pages 5400--5409, 2018.

\bibitem{domain_alignment1}
Ya Li, Xinmei Tian, Mingming Gong, Yajing Liu, Tongliang Liu, Kun Zhang, and Dacheng Tao.
\newblock Deep domain generalization via conditional invariant adversarial networks.
\newblock In {\em Proceedings of the European conference on computer vision (ECCV)}, pages 624--639, 2018.

\bibitem{miyato2018virtual}
Takeru Miyato, Shin-ichi Maeda, Masanori Koyama, and Shin Ishii.
\newblock Virtual adversarial training: a regularization method for supervised and semi-supervised learning.
\newblock {\em IEEE transactions on pattern analysis and machine intelligence}, 41(8):1979--1993, 2018.

\bibitem{muandet2013domain}
Krikamol Muandet, David Balduzzi, and Bernhard Sch{\"o}lkopf.
\newblock Domain generalization via invariant feature representation.
\newblock In {\em ICML}, 2013.

\bibitem{ouali2020semi}
Yassine Ouali, C{\'e}line Hudelot, and Myriam Tami.
\newblock Semi-supervised semantic segmentation with cross-consistency training.
\newblock In {\em Proceedings of the IEEE/CVF Conference on Computer Vision and Pattern Recognition}, pages 12674--12684, 2020.

\bibitem{peng2019moment}
Xingchao Peng, Qinxun Bai, Xide Xia, Zijun Huang, Kate Saenko, and Bo Wang.
\newblock Moment matching for multi-source domain adaptation.
\newblock In {\em ICCV}, pages 1406--1415, 2019.

\bibitem{multimatch}
Lei Qi, Hongpeng Yang, Yinghuan Shi, and Xin Geng.
\newblock Multimatch: Multi-task learning for semi-supervised domain generalization.
\newblock {\em ACM Trans. Multimedia Comput. Commun. Appl.}, 20(6), mar 2024.

\bibitem{recht2019imagenet}
Benjamin Recht, Rebecca Roelofs, Ludwig Schmidt, and Vaishaal Shankar.
\newblock Do imagenet classifiers generalize to imagenet?
\newblock In {\em International conference on machine learning}, pages 5389--5400. PMLR, 2019.

\bibitem{sagawa2019distributionally}
Shiori Sagawa, Pang~Wei Koh, Tatsunori~B Hashimoto, and Percy Liang.
\newblock Distributionally robust neural networks for group shifts: On the importance of regularization for worst-case generalization.
\newblock {\em arXiv preprint arXiv:1911.08731}, 2019.

\bibitem{10.5555/3157096.3157227}
Mehdi Sajjadi, Mehran Javanmardi, and Tolga Tasdizen.
\newblock Regularization with stochastic transformations and perturbations for deep semi-supervised learning.
\newblock In {\em Proceedings of the 30th International Conference on Neural Information Processing Systems}, NIPS'16, page 1171–1179, Red Hook, NY, USA, 2016. Curran Associates Inc.

\bibitem{meta_learning1}
Yang Shu, Zhangjie Cao, Chenyu Wang, Jianmin Wang, and Mingsheng Long.
\newblock Open domain generalization with domain-augmented meta-learning.
\newblock In {\em Proceedings of the IEEE/CVF conference on computer vision and pattern recognition}, pages 9624--9633, 2021.

\bibitem{sohn2020fixmatch}
Kihyuk Sohn, David Berthelot, Nicholas Carlini, Zizhao Zhang, Han Zhang, Colin~A Raffel, Ekin~Dogus Cubuk, Alexey Kurakin, and Chun-Liang Li.
\newblock Fixmatch: Simplifying semi-supervised learning with consistency and confidence.
\newblock {\em Advances in neural information processing systems}, 33:596--608, 2020.

\bibitem{sultana2022self}
Maryam Sultana, Muzammal Naseer, Muhammad~Haris Khan, Salman Khan, and Fahad~Shahbaz Khan.
\newblock Self-distilled vision transformer for domain generalization.
\newblock In {\em ACCV}, pages 3068--3085, 2022.

\bibitem{tarvainen2017mean}
Antti Tarvainen and Harri Valpola.
\newblock Mean teachers are better role models: Weight-averaged consistency targets improve semi-supervised deep learning results.
\newblock {\em Advances in neural information processing systems}, 30, 2017.

\bibitem{tzeng2014deep}
Eric Tzeng, Judy Hoffman, Ning Zhang, Kate Saenko, and Trevor Darrell.
\newblock Deep domain confusion: Maximizing for domain invariance.
\newblock {\em arXiv preprint arXiv:1412.3474}, 2014.

\bibitem{vapnik1999nature}
Vladimir Vapnik.
\newblock {\em The nature of statistical learning theory}.
\newblock Springer science \& business media, 1999.

\bibitem{venkateswara2017deep}
Hemanth Venkateswara, Jose Eusebio, Shayok Chakraborty, and Sethuraman Panchanathan.
\newblock Deep hashing network for unsupervised domain adaptation.
\newblock In {\em CVPR}, pages 5018--5027, 2017.

\bibitem{volpi2018generalizing}
Riccardo Volpi, Hongseok Namkoong, Ozan Sener, John~C Duchi, Vittorio Murino, and Silvio Savarese.
\newblock Generalizing to unseen domains via adversarial data augmentation.
\newblock In {\em NeurIPS}, 2018.

\bibitem{data_augmentation2}
Haotao Wang, Chaowei Xiao, Jean Kossaifi, Zhiding Yu, Anima Anandkumar, and Zhangyang Wang.
\newblock Augmax: Adversarial composition of random augmentations for robust training.
\newblock {\em Advances in neural information processing systems}, 34:237--250, 2021.

\bibitem{wang2023better}
Ruiqi Wang, Lei Qi, Yinghuan Shi, and Yang Gao.
\newblock Better pseudo-label: Joint domain-aware label and dual-classifier for semi-supervised domain generalization.
\newblock {\em Pattern Recognition}, 133:108987, 2023.

\bibitem{wang2020learning}
Shujun Wang, Lequan Yu, Caizi Li, Chi-Wing Fu, and Pheng-Ann Heng.
\newblock Learning from extrinsic and intrinsic supervisions for domain generalization.
\newblock 2020.

\bibitem{wang2023freematch}
Yidong Wang, Hao Chen, Qiang Heng, Wenxin Hou, Yue Fan, , Zhen Wu, Jindong Wang, Marios Savvides, Takahiro Shinozaki, Bhiksha Raj, Bernt Schiele, and Xing Xie.
\newblock Freematch: Self-adaptive thresholding for semi-supervised learning.
\newblock 2023.

\bibitem{xie2020unsupervised}
Qizhe Xie, Zihang Dai, Eduard Hovy, Thang Luong, and Quoc Le.
\newblock Unsupervised data augmentation for consistency training.
\newblock {\em Advances in neural information processing systems}, 33:6256--6268, 2020.

\bibitem{xie2020self}
Qizhe Xie, Minh-Thang Luong, Eduard Hovy, and Quoc~V Le.
\newblock Self-training with noisy student improves imagenet classification.
\newblock In {\em Proceedings of the IEEE/CVF conference on computer vision and pattern recognition}, pages 10687--10698, 2020.

\bibitem{data_augmentation1}
Qinwei Xu, Ruipeng Zhang, Ya Zhang, Yanfeng Wang, and Qi Tian.
\newblock A fourier-based framework for domain generalization.
\newblock In {\em Proceedings of the IEEE/CVF Conference on Computer Vision and Pattern Recognition}, pages 14383--14392, 2021.

\bibitem{yeGraph}
Minxiang Ye, Yifei Zhang, Shiqiang Zhu, Anhuan Xie, and Senwei Xiang.
\newblock Semi-supervised domain generalization with graph-based classifier.
\newblock In {\em ICASSP 2023 - 2023 IEEE International Conference on Acoustics, Speech and Signal Processing (ICASSP)}, pages 1--5, 2023.

\bibitem{yuan2022label}
Junkun Yuan, Xu Ma, Defang Chen, Kun Kuang, Fei Wu, and Lanfen Lin.
\newblock Label-efficient domain generalization via collaborative exploration and generalization.
\newblock In {\em Proceedings of the 30th ACM International Conference on Multimedia}, pages 2361--2370, 2022.

\bibitem{labelefficient}
Junkun Yuan, Xu Ma, Defang Chen, Kun Kuang, Fei Wu, and Lanfen Lin.
\newblock Label-efficient domain generalization via collaborative exploration and generalization.
\newblock In {\em Proceedings of the 30th ACM International Conference on Multimedia}, MM '22, page 2361–2370, New York, NY, USA, 2022. Association for Computing Machinery.

\bibitem{zhang2021flexmatch}
Bowen Zhang, Yidong Wang, Wenxin Hou, Hao Wu, Jindong Wang, Manabu Okumura, and Takahiro Shinozaki.
\newblock Flexmatch: Boosting semi-supervised learning with curriculum pseudo labeling.
\newblock {\em Advances in Neural Information Processing Systems}, 34:18408--18419, 2021.

\bibitem{zhang2018mixup}
Hongyi Zhang, Moustapha Cisse, Yann~N Dauphin, and David Lopez-Paz.
\newblock mixup: Beyond empirical risk minimization.
\newblock {\em In ICLR (ICLR)}, 2018.

\bibitem{zhang2023semisupervised}
Lei Zhang, Ji-Fu Li, and Wei Wang.
\newblock Semi-supervised domain generalization with known and unknown classes.
\newblock In {\em Thirty-seventh Conference on Neural Information Processing Systems}, 2023.

\bibitem{domain_alignment2}
Shanshan Zhao, Mingming Gong, Tongliang Liu, Huan Fu, and Dacheng Tao.
\newblock Domain generalization via entropy regularization.
\newblock {\em Advances in Neural Information Processing Systems}, 33:16096--16107, 2020.

\bibitem{data_augmentation3}
Zhun Zhong, Yuyang Zhao, Gim~Hee Lee, and Nicu Sebe.
\newblock Adversarial style augmentation for domain generalized urban-scene segmentation.
\newblock {\em Advances in Neural Information Processing Systems}, 35:338--350, 2022.

\bibitem{zhou2022survey}
Kaiyang Zhou, Ziwei Liu, Yu Qiao, Tao Xiang, and Chen~Change Loy.
\newblock Domain generalization: A survey.
\newblock {\em IEEE Transactions on Pattern Analysis and Machine Intelligence}, 2022.

\bibitem{zhou2023semi}
Kaiyang Zhou, Chen~Change Loy, and Ziwei Liu.
\newblock Semi-supervised domain generalization with stochastic stylematch.
\newblock {\em International Journal of Computer Vision}, pages 1--11, 2023.

\bibitem{zhou2020deep}
Kaiyang Zhou, Yongxin Yang, Timothy Hospedales, and Tao Xiang.
\newblock Deep domain-adversarial image generation for domain generalisation.
\newblock In {\em Proceedings of the AAAI Conference on Artificial Intelligence}, volume~34, pages 13025--13032, 2020.

\bibitem{zhou2020learning}
Kaiyang Zhou, Yongxin Yang, Timothy Hospedales, and Tao Xiang.
\newblock Learning to generate novel domains for domain generalization.
\newblock In {\em European Conference on Computer Vision}, pages 561--578. Springer, 2020.

\end{thebibliography}
}
\clearpage
\pagebreak 


\appendix

\section{Detailed description of datasets}
\label{section:Detailed description of datasets}
We conduct experiments on six challenging and diverse DG datasets to validate the effectiveness of the proposed method. \textbf{PACS} \cite{li2017deeper} contains 7 categories of images from four domains (Photo, Art painting, Cartoon, and Sketch). 
%
\textbf{OfficeHome} \cite{venkateswara2017deep} consists of images from four different domains (Art, Clipart, Product, and Real-world). It encompasses 65 object categories that are commonly encountered in office and home environments. 
\textbf{VLCS} \cite{fang2013unbiased} comprises images spanning across four domains with 5 categories and has four domains (Caltech, Labelme, SUN, and Pascal). \textbf{Digits-DG} \cite{volpi2018generalizing} includes digit images drawn from MNIST, SVHN, MNIST-M and SYNTH. \textbf{Terra Incognita} \cite{beery2018recognition} contains photos of wild animals taken by cameras at different locations (location 38, location 43, location 46, and location 100) with 10 classes. \textbf{DomainNet} \cite{peng2019moment} is a large-scale dataset of common objects in six different domains (clipart, infograph, real, painting, quickdraw, sketch) with 345 categories of objects.

\section{Impact of noise perturbation injected}
\label{section:Impact of noise perturbation injected}

We vary the variance $\e^2$ of the isotropic Gaussian $\cN(0, \e^2I)$ and evaluate the impact on our method in Tab.~\ref{tab:noise}. We note that the performance of the method is mostly insensitive to variances 0.1, 0.5, and 1.0. The best performance is achieved at 1.0, however, it decreases upon doubling the variance to 2.0.

\begin{table}[h]

\centering
\scalebox{0.9}{
\begin{tabular}{cc}
\toprule
$\e^2$                                                                                     & Average \\ \midrule
0.1 & 59.5     \\
0.5 &  59.3 \\
1.0  & \textbf{59.7} \\
2.0 & 57.3 \\
   \bottomrule   
\end{tabular}}
\caption{Results with different values of variance $\e^2$ of the isotropic Gaussian $\cN(0, \e^2I)$. Results are shown for the OfficeHome dataset under 10 labels setting.}
\label{tab:noise}

\end{table}

\section{Pseudo-labeling accuracy vs. Confidence threshold}
\label{section:Pseudo-labeling accuracy vs. Confidence threshold}
\begin{figure}[h]
    \centering
     \includegraphics[width=1.0\linewidth]{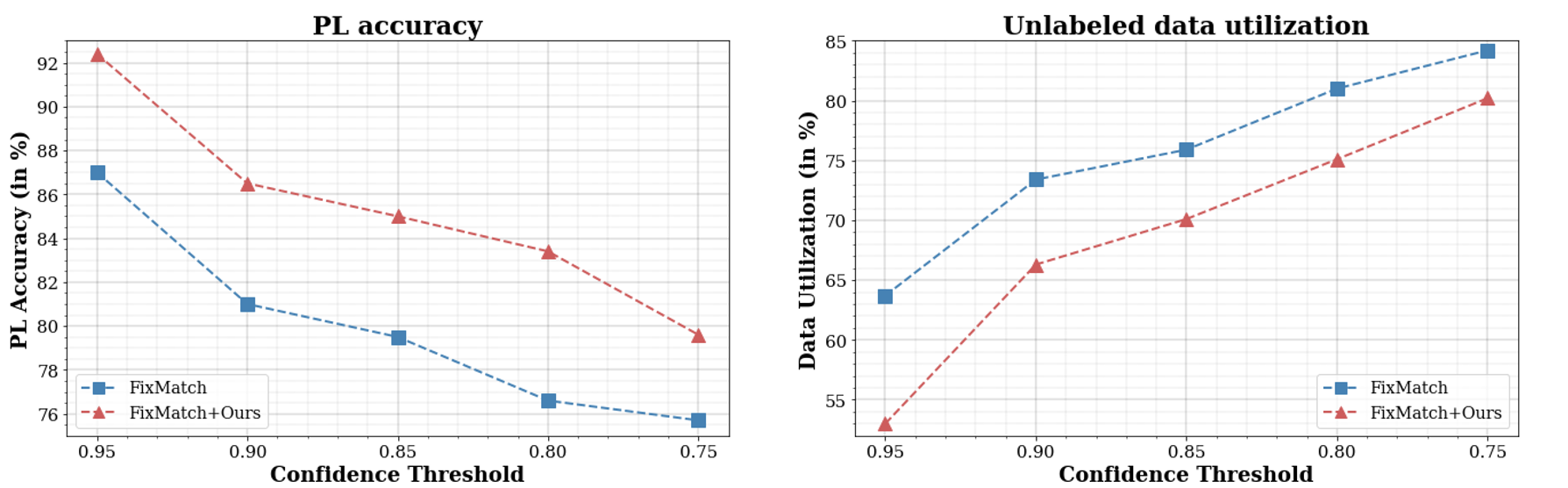}
    \caption{(Left) Pseudo-label accuracy upon varying the confidence threshold in FixMatch and our method. (Right) Unlabelled data utilization i.e. the percentage of unlabelled data that passes the confidence threshold for both FixMatch and our method. These results are shown on the OfficeHome dataset with 10 label settings.}
    \label{fig:PL_vs_threshold}
\end{figure}

Fig.~\ref{fig:PL_vs_threshold} (left) shows the variation of pseudo-labeling accuracy against the confidence threshold \cite{sohn2020fixmatch} on the OfficeHome  dataset under the 10 labels setting for FixMatch \cite{sohn2020fixmatch} and our method. Our proposed method retains a higher pseudo-labeling accuracy than the baseline \cite{sohn2020fixmatch} even when we lower the confidence threshold. Furthermore, we plot the unlabeled data utilization i.e. the percentage of unlabelled data  that passes the confidence threshold for both FixMatch and our method as the confidence threshold varies (see Fig.~\ref{fig:PL_vs_threshold} (right)). The weight modulation technique in our method tends to reduce the model’s maximum confidence when computing pseudo-labels. As a result, only highly accurate pseudo-labels will make it past the threshold.

 \section{Comparison with DG baselines}
\label{section:Comparison with DG baselines}

We show the performance of several DG methods: (ERM \cite{vapnik1999nature}, MixUp \cite{zhang2018mixup}, and GroupDRO \cite{sagawa2019distributionally} and also show results after combining these DG methods and pseudo-labelling from FixMatch \cite{sohn2020fixmatch}. Tab.~\ref{tab:setting1_DG} and Tab.~\ref{tab:one_fully_labeled_DG} report results with the first SSDG setting and the second SSDG setting, respectively.

\begin{table*}[]
  \centering

  \resizebox{\textwidth}{!}{
  \setlength{\tabcolsep}{0.9mm}{
  \begin{tabular}{c|cccccc|cccccc}
    \toprule
    \multirow{2}{*}{\textbf{Method}}   &   \multicolumn{6}{c|}{\textbf{5 labels}}  &   \multicolumn{6}{c}{\textbf{10 labels}}   \\
    ~   &   \textbf{PACS}  &   \textbf{OfficeHome}    &   \textbf{VLCS}    &   \textbf{DigitsDG}  &   \textbf{TerraInc}  &   \textbf{DomainNet}   &   \textbf{PACS}  &   \textbf{OfficeHome}    &   \textbf{VLCS}    &   \textbf{DigitsDG}  &   \textbf{TerraInc}  &   \textbf{DomainNet}   \\ \midrule
    
    ERM &   51.2$_{\pm3.0}$    &   51.7$_{\pm0.6}$    &   67.2$_{\pm1.8}$    &   22.7$_{\pm1.0}$    &   22.9$_{\pm3.0}$    &   23.5$_{\pm0.2}$  &  
    
    59.8$_{\pm2.3}$   &   56.7$_{\pm0.8}$    &   68.0$_{\pm0.3}$    &   29.1$_{\pm2.9}$    &   23.5$_{\pm1.2}$    &   29.4$_{\pm0.1}$ \\
    
    MixUp &   45.3$_{\pm3.8}$    &   52.7$_{\pm0.6}$    &   69.9$_{\pm1.3}$    &      21.7$_{\pm1.9}$   &   21.0$_{\pm2.9}$    &   23.5$_{\pm0.3}$ &  
    
    58.5$_{\pm2.2}$    &   57.2$_{\pm0.6}$    &   69.6$_{\pm1.0}$    &   29.7$_{\pm3.1}$    &   24.8$_{\pm3.3}$   &   28.8$_{\pm0.1}$ \\
    
    GroupDRO &   48.2$_{\pm3.6}$    &   53.8$_{\pm0.6}$    &   69.8$_{\pm1.2}$    &   23.1$_{\pm1.9}$    &   22.4$_{\pm3.1}$    &   20.2$_{\pm0.2}$    &  
    
    57.3$_{\pm1.2}$    &   57.8$_{\pm0.4}$    &   69.4$_{\pm0.9}$    &   31.5$_{\pm2.5}$    &   25.8$_{\pm3.3}$    &   26.5$_{\pm0.5}$\\
    
    \midrule
    
    ERM + PL &   62.8$_{\pm3.0}$    &   54.2$_{\pm0.6}$    &   65.4$_{\pm2.9}$    &   43.4$_{\pm2.9}$    &   25.4$_{\pm3.2}$    &   24.1$_{\pm0.2}$   &   
    
    63.0$_{\pm1.5}$    &   55.5$_{\pm0.3}$    &   60.5$_{\pm1.1}$    &   55.0$_{\pm2.4}$    &   26.8$_{\pm1.5}$ &   26.7$_{\pm0.1}$   \\
    
    MixUp + PL &   60.6$_{\pm2.9}$    &   51.9$_{\pm0.4}$    &   60.8$_{\pm2.8}$    &   35.4$_{\pm1.3}$    &   24.1$_{\pm3.0}$    &   23.3$_{\pm0.2}$ &   
    
    62.3$_{\pm1.9}$   &   55.1$_{\pm0.2}$    &   64.4$_{\pm1.1}$    &   43.5$_{\pm1.0}$    &    27.6$_{\pm2.2}$ &   28.5$_{\pm0.3}$\\
    
    GroupDRO + PL &   62.3$_{\pm1.9}$    &   54.5$_{\pm0.5}$    &   69.3$_{\pm0.3}$    &   39.4$_{\pm1.3}$    &   25.1$_{\pm3.2}$    &   25.6$_{\pm0.2}$    &   
    
    62.1$_{\pm2.0}$    &   58.5$_{\pm0.3}$    &   66.5$_{\pm0.2}$    &   49.9$_{\pm1.9}$    &   26.9$_{\pm1.2}$    &   28.0$_{\pm0.1}$\\

    \bottomrule
    \end{tabular}
    
    }
}   
   \caption{Comparison with the DG methods, DG+PL\cite{sohn2020fixmatch} methods under the first setting i.e only a few instances({5,10}) are labeled from each source domain.  
}
  
 \label{tab:setting1_DG}
    \end{table*}

\begin{table*}[]
  \centering

  \scalebox{0.77}{
  \setlength{\tabcolsep}{3mm}{
  \begin{tabular}{c|cccccc}
    \toprule
    \textbf{Method}  &   \textbf{PACS}  &   \textbf{OfficeHome}    &   \textbf{VLCS}    &   \textbf{Digits}  &   \textbf{TerraInc}  &   \textbf{DomainNet}    \\\midrule
    ERM &   69.8$\pm$1.8    &   61.7$\pm$0.4    &   60.8$\pm$0.7    &   36.7$\pm$0.7    &   40.0$\pm$2.3    &   33.1$\pm$0.1\\
    MixUp &   66.9$\pm$1.9    &   61.6$\pm$0.2    &   61.3$\pm$0.5    &   40.1$\pm$1.0    &   40.1$\pm$0.8    &   33.9$\pm$0.1\\
    GroupDRO &   71.6$\pm$1.3    &   63.7$\pm$0.1    &   61.5$\pm$0.7    &   38.8$\pm$0.7    &   40.5$\pm$1.3&   34.1$\pm$0.1\\
    \midrule
    ERM+PL &   65.2$\pm$1.6    &   60.4$\pm$0.4    &   50.5$\pm$0.8    &   53.4$\pm$0.9    &   41.1$\pm$0.8    &   31.4$\pm$0.1\\
    MixUp+PL &   66.9$\pm$1.4    &   62.0$\pm$0.3    &   55.9$\pm$0.4    &   49.3$\pm$0.3    &   38.2$\pm$1.3    &   35.5$\pm$0.2\\
    GroupDRO+PL &   78.6$\pm$1.9   &   64.5$\pm$0.1    &   55.8$\pm$0.6    &   40.5$\pm$1.0    &   42.5$\pm$0.4    &   35.1$\pm$0.1\\
    \bottomrule
    \end{tabular} }
    }
      \caption{Comparison with the DG methods, DG+PL\cite{sohn2020fixmatch} methods under the first setting i.e one source domain is completely labeled and the other completely unlabeled. 
    }
  \label{tab:one_fully_labeled_DG}
\end{table*}

\section{Performance under class-imbalance } 
VLCS \cite{fang2013unbiased} has a significant class imbalance than most of the DG datasets. In Tab.~\ref{tab:imbalance} we calculate the ratio between the number of samples for the highest and lowest available classes in each domain. It should be noted that our proposed method shows notable gains of $+5.3\%$ and $+5.2\%$ for 5 and 10 labels settings respectively.

\begin{table}[h]
    \centering
    \scalebox{0.8}{
    \begin{tabular}{cccc}
            \toprule[0.15em]
            \multirow{2}{*}{Domain} & \multicolumn{2}{c}{VLCS \# of samples} & \multirow{2}{*}{Ratio} \\
               ~   &   Highest &   Lowest  &   ~ \\
            \midrule
           Caltech &   809 &   62  &   13.0   \\
         LabelMe &   1124 &   39  &   28.9   \\
            Pascal &   1394 &   307  &   4.6   \\
            SUN &   1175 &   19  &   61.9   \\
            \bottomrule
        \end{tabular}}
    \caption{Num. of samples for highest and lowest available classes for each domain.}
    \label{tab:imbalance}
\end{table}

\section{Architectural details of encoder-decoder-like pair}
\label{section:en-dec}

For the encoder, we use 3 linear layers each followed by a ReLU\cite{Agarap2018DeepLU} activation layer, and reduce the size of the embedding dimension by a factor of 2. Intermediate embedding concatenated with noise will follow a two-linear layer decoder each followed by a ReLU\cite{Agarap2018DeepLU} activation layer.

\section{t-SNE visualization of domain information vector}
\begin{figure}[h]
    \centering
     \includegraphics[width=.7\linewidth]{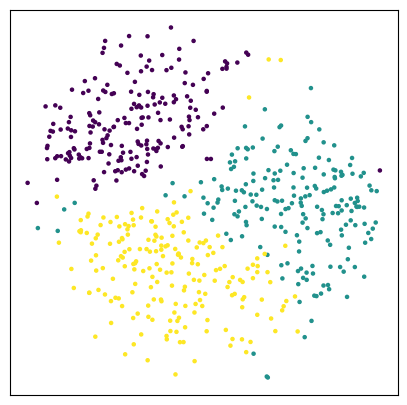}
    \caption{t-SNE visualization of domain information vectors $I^k$ taken during training on OfficeHome dataset.}
    \label{fig:tsne}
\end{figure}

The mini-batch mean is a simple way of aggregating the domain-specific information \cite{dumoulin2017a, Huang2017ArbitraryST} as samples in the same mini-batch are drawn from the same domain. t-SNE visualization (see Fig.~\ref{fig:tsne}) of domain information vectors $I^k$ taken during training indicates that these domain information vectors are distinct for each source domain (3 source domains). 

\section{Additional details on the second setting}
\label{section:Additional details on the second setting}
Under the second SSDG setting, for a given dataset, we select a target domain and keep it fixed while making each source domain labeled and others unlabeled and report the average recognition accuracy. The fixed target domain in each dataset is as follows: ”Photo” in PACS, ”Real-world” in OfficeHome, ”SUN” in VLCS, ”SVHN” in DigitsDG, ”location 100” in TerraIncognita and ”Real” in DomainNet. Each experiment is conducted for 5 independent trials.

\end{document}